\documentclass{article}
\usepackage{spconf,amsmath,graphicx,booktabs}

\title{Transsion TSUP's speech recognition system for ASRU 2023 MADASR Challenge}
\name{Xiaoxiao Li\textsuperscript{*}, Gaosheng Zhang\textsuperscript{*}, An Zhu, Weiyong Li, Shuming Fang, Xiaoyue Yang, Jianchao Zhu\thanks{\textsuperscript{*}Equal contribution}}
\address{Shenzhen Transsion Holdings Co., Ltd, Shanghai Branch, China \\
\small \{xiaoxiao.li7, gaosheng.zhang, an.zhu, weiyong.li, shuming.fang, xiaoyue.yang, jianchao.zhu5\}@transsion.com}
\begin{document}
%
\maketitle
\begin{abstract}
This paper presents a speech recognition system developed by the Transsion Speech Understanding Processing Team (TSUP) for the ASRU 2023 MADASR Challenge. The system focuses on adapting ASR models for low-resource Indian languages and covers all four tracks of the challenge. For tracks 1 and 2, the acoustic model utilized a squeezeformer encoder and bidirectional transformer decoder with joint CTC-Attention training loss. Additionally, an external KenLM language model was used during TLG beam search decoding. For tracks 3 and 4, pretrained IndicWhisper models were employed and finetuned on both the challenge dataset and publicly available datasets. The whisper beam search decoding was also modified to support an external KenLM language model, which enabled better utilization of the additional text provided by the challenge. The proposed method achieved word error rates (WER) of 24.17\%, 24.43\%, 15.97\%, and 15.97\% for Bengali language in the four tracks, and WER of 19.61\%, 19.54\%, 15.48\%, and 15.48\% for Bhojpuri language in the four tracks. These results demonstrate the effectiveness of the proposed method.
\end{abstract}
\section{Introduction}
\label{sec:intro}

The ASRU 2023 MADASR Challenge \footnote{https://sites.google.com/view/respinasrchallenge2023/home} \cite{singh2023model} aims to adapt ASR models for low-resource Indian languages, specifically Bengali and Bhojpuri. The challenge provides approximately 850 hours of read-speech data spoken by nearly 2000 speakers. Participants are encouraged to explore the importance of training data and modeling techniques such as Whisper \cite{radford2023robust} and wav2vec 2.0 \cite{baevski2020wav2vec}. The evaluation metrics used on the blind test dataset are the word and character error rates (WER and CER). There are four different tracks with varying degrees of data restrictions:
\begin{itemize}
\item Track 1 only allows using the data shared through the special session (in-corpus) for both acoustic and text.
\item Track 2 allows using in-corpus acoustic data and any text data.
\item Track 3 allows using any acoustic data including pretrained model and in-corpus text data.
\item Track 4 encourages participants to use any existing resources along with the in-corpus audio and text to get the best performance.
\end{itemize}
This paper describes the speech recognition system developed by the Transsion Speech Understanding Processing Team (TSUP) for the Challenge covering all the four tracks. In tracks 1 and 2, the focus is on model architectures and units, while in tracks 3 and 4, the focus is on leveraging pretrained models and fine-tuning.  

\section{DATA}
\label{sec:intro}

The challenge offers approximately 850 hours of read-speech data spoken by nearly 2000 speakers for both Bengali and Bhojpuri. In addition, dialect-rich text corpora composed by language experts of the corresponding dialects are also included. These in-corpus datasets are part of the RESPIN project, which aims to collect dialect-rich read-speech corpora in 9 Indian languages and can be used in all four tracks. Furthermore, publicly available datasets for both Bengali and Bhojpuri were also collected and are used in track3 and track4 only. No additional text corpora was investigated in our experiment. Table 1 listed the speech and text datasets information.

\begin{table}[h]
    \centering
    \caption{Speech and text data summary.}
    \label{tab:univ-compa}
    \setlength{\tabcolsep}{0.1mm}
    \begin{tabular}{ccc}
    \toprule
        \textbf{ } & \textbf{Bengali\ } & \textbf{\ Bhojpuri}  \\ \midrule
        in-corpus datasets  \\ \midrule        
        challenge speech data(\#Hours) & 852 & 838  \\ 
        dialect-rich text corpora(\#Sentences) & 194K & 228K  \\ \midrule
        publicly available datasets  \\ \midrule
        bengali\_SLR53(\#Hours) & 229  \\
        bengali\_SLR37(\#Hours) & 3 &  \\
        common\_voice\_13(\#Hours) & 1244.33 &  \\
        kathbath\_bengali(\#Hours) & 123.03 &  \\
        shrutilipi\_bengali(\#Hours) & 440.33 &  \\
        ULCA-asr-dataset-corpus(\#Hours) &  & 60 \\
    \bottomrule
    \end{tabular}
\end{table}

\section{Track 3 and Track 4 System}
\label{sec:format}

Recently, the Vistaar \cite{bhogale2023vistaar} project released IndicWhisper models, which cover 12 Indian languages including Bengali and Hindi. However, Bhojpuri was not included in these models. We hypothesize that due to the linguistic similarity between Hindi and Bhojpuri, it may be possible to use Hindi models to decode Bhojpuri audio. Table 2 shows the zero-shot performance of IndicWhisper models on the challenge dev datasets, indicating that Bengali and Hindi IndicWhisper models demonstrate satisfactory zero-shot performance. Our focus has been on fine-tuning these two models for track 3 and 4.

\begin{table}[h]
    \centering
    \caption{IndicWhisper zero-shot performance on Bengali and Bhojpuri dev dataset (WER).}
    \label{tab:univ-compa}
    \setlength{\tabcolsep}{0.1mm}
    \begin{tabular}{ccc}
    \toprule
        \textbf{ } & \textbf{Bengali dev } & \textbf{\ Bhojpuri dev}  \\ \midrule
        Bengali IndicWhisper & 34.1\%  &   \\ 
        Hindi IndicWhisper &  & 50.3\%  \\ 
     
    \bottomrule
    \end{tabular}
\end{table}

\subsection{IndicWhisper fine-tuning}
\label{ssec:subhead}

The IndicWhisper model follows the same structure as openai's whisper-medium, with each model containing 769M parameters and 24 layers in both the encoder and decoder. To fine-tune all layers of the models, we used four A100 GPUs with 40GB GPU RAM, with a learning rate of 5e-6 and linearly decay to zero after 3 epochs. However, we found that the fine-tuning process easily led to overfitting due to the large model size and relatively small dataset. Therefore, it's important to apply dropout of 0.1 during the fine-tuning process to avoid overfitting. Finally, we selected the best checkpoint based on the WER of the dev dataset. Table 3 shows the fine-tuning results of IndicWhisper models on different datasets.

\begin{table}[h]
    \centering
    \caption{fine-tuning results of IndicWhisper models.}
    \label{tab:univ-compa}
    \setlength{\tabcolsep}{0.1mm}
    \begin{tabular}{ccc}
    \toprule
        \textbf{ } & \textbf{Bengali dev\ } & \textbf{\ Bhojpuri dev}  \\ \midrule
        Bengali \\ \midrule
        track 3 challenge baseline  & 19.1\%  &   \\ 
        track 4 challenge baseline  & 15.71\%  &   \\ 
        zero-shot  & 34.1\%  &   \\ 
        fine-tuning on challenge speech data & 17.50\%  &   \\
        fine-tuning on all datasets & 16.81\%  &   \\ 
        decoding with LM & \textbf{14.85\%}  &   \\ \midrule
        Bhojpuri \\ \midrule
        track 3 challenge baseline & & 18.13\%   \\ 
        track 4 challenge baseline & & 15.7\%   \\ 
        zero-shot & & 50.3\%     \\ 
        fine-tuning on challenge speech data &  & 16.97\%   \\
        fine-tuning on all datasets &  & 16.25\%   \\
        decoding with LM  & & \textbf{14.19\%}   \\
    \bottomrule
    \end{tabular}
\end{table}

\subsection{Language Model Decoding}
\label{ssec:subhead}

The original whisper decoding is based on beam search and doesn’t support the external language model. To improve the utilization of the dialect-rich text corpora provided by the challenge, the decoding code was modified and the BPE-level kenlm language model was fused. A 20-gram BPE-level kenlm language model was trained on the challenge speech data text and the dialect-rich text in our experiment. Optimized language model weights were determined based on the lowest WER on the dev dataset. The results from Table 3 showed that the external language model brought a large margin WER improvement, with a 1.96\% absolute reduction for Bengali and a 2.06\% absolute reduction for Bhojpuri.

\subsection{Final system}
\label{ssec:subhead}

We submitted the same system to track 3 and track 4 which was based on the fine-tuning model on all speech datasets and decoding with LM. We achieved wer of 15.97\% and 15.48\% on Bengali and Bhojpuri blind test dataset, which show a large improvements over the provided baseline(Table 4).

\begin{table}[h]
    \centering
    \caption{Final submission system for track 3 and track 4.}
    \label{tab:univ-compa}
    \setlength{\tabcolsep}{0.1mm}
    \begin{tabular}{ccc}
    \toprule
        \textbf{ } & \textbf{Bengali test\ } & \textbf{\ Bhojpuri test}  \\ \midrule
        Bengali \\ \midrule
        track 3 challenge baseline  & 20.69\%  &   \\ 
        track 4 challenge baseline  & 18.46\%  &   \\ 
        our submission & \textbf{15.97\%}  &   \\ \midrule
        Bhojpuri \\ \midrule
        track 3 challenge baseline & & 18.74\%   \\ 
        track 4 challenge baseline & & 16.99\%   \\ 
         our submission & & \textbf{15.48\%}   \\
    \bottomrule
    \end{tabular}
\end{table}

\bibliographystyle{IEEEbib}
\bibliography{refs}

\begin{thebibliography}{1}

\bibitem{singh2023model}
Abhayjeet Singh, Arjun~Singh Mehta, Ashish Khuraishi~K S, Deekshitha G, Gauri
  Date, Jai Nanavati, Jesuraja Bandekar, Karnalius Basumatary, Karthika P,
  Sandhya Badiger, Sathvik Udupa, Saurabh Kumar, Savitha, Prasanta~Kumar Ghosh,
  Prashanthi V, Priyanka Pai, Raoul Nanavati, Rohan Saxena, Sai Praneeth~Reddy
  Mora, and Srinivasa Raghavan,
\newblock ``Model adaptation for asr in low-resource indian languages,'' 2023.

\bibitem{radford2023robust}
Alec Radford, Jong~Wook Kim, Tao Xu, Greg Brockman, Christine McLeavey, and
  Ilya Sutskever,
\newblock ``Robust speech recognition via large-scale weak supervision,''
\newblock in {\em International Conference on Machine Learning}. PMLR, 2023,
  pp. 28492--28518.

\bibitem{baevski2020wav2vec}
Alexei Baevski, Yuhao Zhou, Abdelrahman Mohamed, and Michael Auli,
\newblock ``wav2vec 2.0: A framework for self-supervised learning of speech
  representations,''
\newblock {\em Advances in neural information processing systems}, vol. 33, pp.
  12449--12460, 2020.

\bibitem{bhogale2023vistaar}
Kaushal~Santosh Bhogale, Sai Sundaresan, Abhigyan Raman, Tahir Javed, Mitesh~M
  Khapra, and Pratyush Kumar,
\newblock ``Vistaar: Diverse benchmarks and training sets for indian language
  asr,''
\newblock {\em arXiv preprint arXiv:2305.15386}, 2023.

\end{thebibliography}

\end{document}